\documentclass[aip,cha,reprint,superscriptaddress, floatfix]{revtex4-1}
\usepackage{amsmath}
\usepackage{amsfonts}
\usepackage{amssymb}
\usepackage{graphicx}
\usepackage{xcolor}
\newcommand{\vv}[1]{{\bf #1}}

\begin{document}
\title{Hybrid Forecasting of Chaotic Processes: Using Machine Learning in Conjunction with a Knowledge-Based Model}

\author{Jaideep Pathak}
\affiliation{University of Maryland, College Park, Maryland 20742, USA.}
\author{Alexander Wikner}
\affiliation{Rice University, Houston, Texas 77005, USA.}
\author{Rebeckah Fussell}
\affiliation{Haverford College, Haverford, Pennsylvania 19041, USA.}
\author{Sarthak Chandra}
\affiliation{University of Maryland, College Park, Maryland 20742, USA.}
\author{Brian R. Hunt}
\affiliation{University of Maryland, College Park, Maryland 20742, USA.}
\author{Michelle Girvan}
\affiliation{University of Maryland, College Park, Maryland 20742, USA.}
\author{Edward Ott}
\affiliation{University of Maryland, College Park, Maryland 20742, USA.}




\date{\today}

\begin{abstract}
A model-based approach to forecasting chaotic dynamical systems utilizes knowledge of the physical processes governing the dynamics to build an approximate mathematical model of the system. In contrast, machine learning techniques have demonstrated promising results for forecasting chaotic systems purely from past time series measurements of system state variables (training data), without prior knowledge of the system dynamics. The motivation for this paper is the potential of machine learning for filling in the gaps in our underlying mechanistic knowledge that cause widely-used knowledge-based models to be inaccurate. Thus we here propose a general method that leverages the advantages of these two approaches by combining a knowledge-based model and a machine learning technique to build a hybrid forecasting scheme. Potential applications for such an approach are numerous (e.g., improving weather forecasting). We demonstrate and test the utility of this approach using a particular illustrative version of a machine learning known as reservoir computing, and we apply the resulting hybrid forecaster to a low-dimensional chaotic system, as well as to a high-dimensional spatiotemporal chaotic system. These tests yield extremely promising results in that our hybrid technique is able to accurately predict for a much longer period of time than either its machine-learning component or its model-based component alone.
\end{abstract}


\maketitle

\textbf{Prediction of dynamical system states (e.g., as in weather forecasting) is a common and essential task with many applications in science and technology. This task is often carried out via a system of dynamical equations derived to model the process to be predicted. Due to deficiencies in knowledge or computational capacity, application of these models will generally be imperfect and may give unacceptably inaccurate results. On the other hand data-driven methods, independent of derived knowledge of the system, can be computationally intensive and require unreasonably large amounts of data. In this paper we consider a particular hybridization technique for combining these two approaches. Our tests of this hybrid technique suggest that it can be extremely effective and widely applicable.}

\section{Introduction}\label{sec:introduction}

One approach to forecasting the state of a dynamical system starts by using whatever knowledge and understanding is available about the mechanisms governing the dynamics to construct a mathematical model of the system. Following that, one can use measurements of the system state to estimate initial conditions for the model that can then be integrated forward in time to produce forecasts. We refer to this approach as knowledge-based prediction. The accuracy of knowledge-based prediction is limited by any errors in the mathematical model. Another approach that has recently proven effective is to use machine learning to construct a model purely from extensive past measurements of the system state evolution (training data). Because the latter approach typically makes little or no use of mechanistic understanding, the amount of training data and the computational resources required can be prohibitive, especially when the system to be predicted is large and complex. The purpose of this paper is to propose, describe, and test a general framework for combining a knowledge-based approach with a machine learning approach to build a hybrid prediction scheme with significantly enhanced potential for performance and feasibility of implementation as compared to either an approximate knowledge-based model acting alone or a machine learning model acting alone. The results of our tests of our proposed hybrid scheme suggest that it can have wide applicability for forecasting in many areas of science and technology. We note that hybrids of machine learning with other approaches have previously been applied to a variety of other tasks, but here we consider the general problem of forecasting a dynamical system with an imperfect knowledge-based model, the form of whose imperfections is unknown. Examples of such other tasks addressed by machine learning hybrids include network anomaly detection~\cite{shon}, credit rating~\cite{tsai}, and chemical process modeling~\cite{ungar}, among others.

Another view motivating our hybrid approach is that, when trying to predict the evolution of a system, one might intuitively expect the best results when making appropriate use of \textit{all} the available information about the system. Here we think of both the (perhaps imperfect) physical model and the past measurement data as being two types of system information which we wish to simultaneously and efficiently utilize. The hybrid technique proposed in this paper does this.

To illustrate the hybrid scheme, we focus on a particular type of machine learning known as `reservoir computing'~\cite{jaeger2001echo, maass2002real, jaeger2009reservoir}, which has been previously applied to the prediction of low dimensional systems~\cite{jaeger2004harnessing} and, more recently, to the prediction of large spatiotemporal chaotic systems \cite{pathak2017using, pathakmodel}. We emphasize that, while our illustration is for reservoir computing with a reservoir based on an artificial neural network, we view the results as a general test of the hybrid approach. As such, these results should be relevant to other versions of machine learning~\cite{goodfellow2016deep} (such as Long Short-Term Memory networks~\cite{hochreiter1997long}), as well as reservoir computers in which the reservoir is implemented by various physical means (e.g., electro-optical schemes~\cite{larger2012photonic, larger2017high, antonik2017brain} or Field Programmable Gate Arrays~\citep{haynes2015reservoir}) A particularly dramatic example illustrating the effectiveness of the hybrid approach is shown in Figs.~\ref{fig:ksplots}(d,e,f) in which, when acting alone, both the knowledge-based predictor and the reservoir machine learning predictor give fairly worthless results (prediction time of only a fraction of a Lyapunov time), but, when the same two systems are combined in the hybrid scheme, good predictions are obtained for a substantial duration of about $4$ Lyapunov times. (By a `Lyapunov time' we mean the typical time required for an $e$-fold increase of the distance between two initially close chaotic orbits; see Sec.~\ref{sec:lorenz} and \ref{sec:ks}.)

The rest of this paper is organized as follows. In Sec.~\ref{sec:methods}, we provide an overview of our methods for prediction by using a knowledge-based model and for prediction by exclusively using a reservoir computing model (henceforth referred to as the reservoir-only model). We then describe the hybrid scheme that combines the knowledge-based model with the reservoir-only model. In Sec.~\ref{sec:implementation}, we describe our specific implementation of the reservoir computing scheme and the proposed hybrid scheme using a recurrent-neural-network implementation of the reservoir computer. In Sec.~\ref{sec:lorenz} and \ref{sec:ks}, we demonstrate our hybrid prediction approach using two examples, namely, the low-dimensional Lorenz system~\cite{lorenz1963deterministic} and the high dimensional, spatiotemporal chaotic Kuramoto-Sivashinsky system~\cite{kuramoto1976persistent,sivashinsky1977nonlinear}.


\section{Prediction Methods}\label{sec:methods}

\begin{figure}[h]
	\centering
	\includegraphics[width=0.5\textwidth]{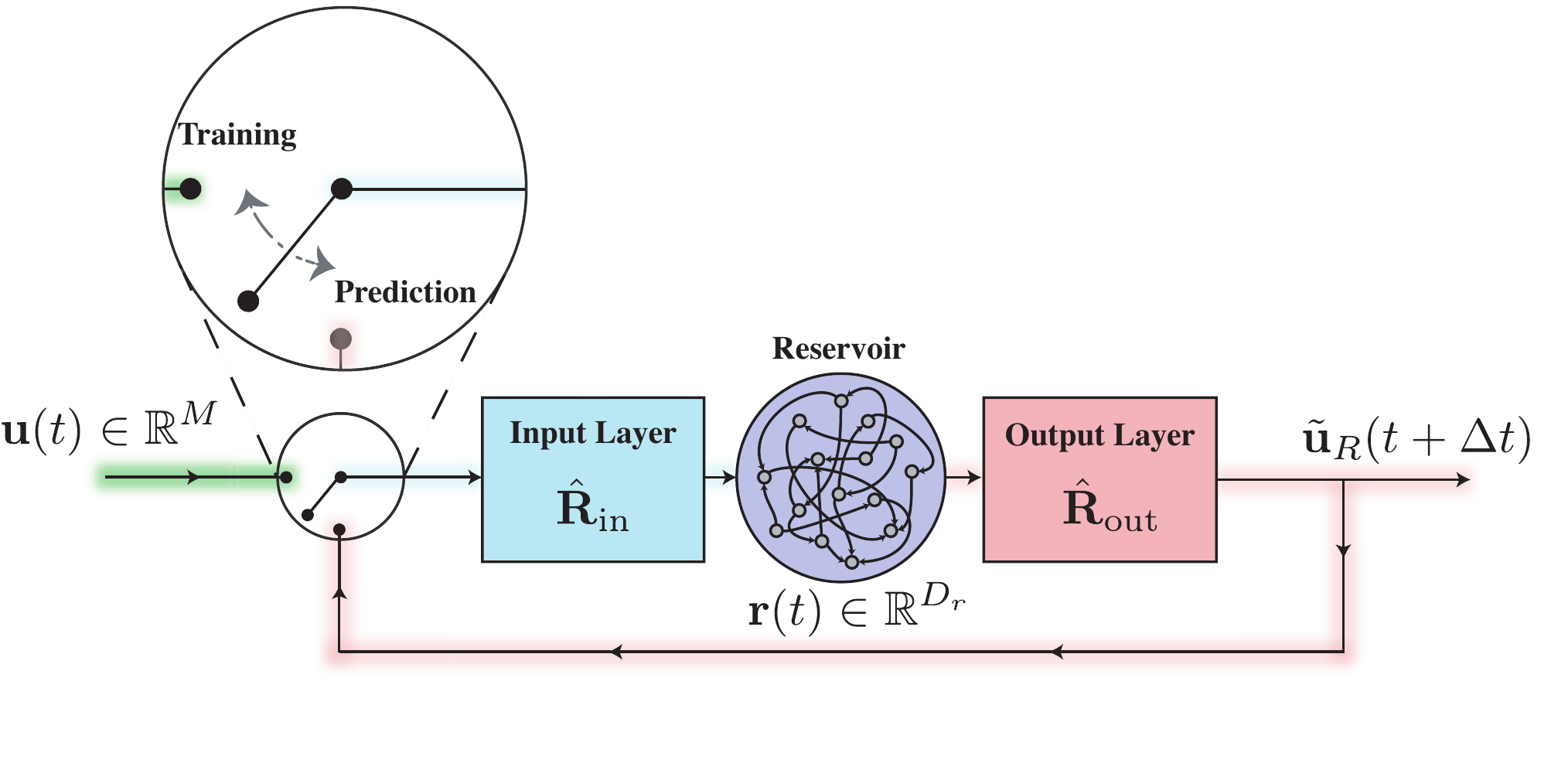}
	\caption{Schematic diagram of reservoir-only prediction setup.}
	\label{fig:res_diagram}
\end{figure}

We consider a dynamical system for which there is available a time series of a set of $M$ measurable state-dependent quantities, which we represent as the $M$ dimensional vector $\mathbf{u}(t)$. As discussed earlier, we propose a hybrid scheme to make predictions of the dynamics of the system by combining an approximate knowledge-based prediction via an approximate model with a purely data-driven prediction scheme that uses machine learning. We will compare predictions made using our hybrid scheme with the predictions of the approximate knowledge-based model alone and predictions made by exclusively using the reservoir computing model.

\subsection{Knowledge-Based Model}\label{sec:sub_model}
We obtain predictions from the approximate knowledge-based model acting alone assuming that the knowledge-based model is capable of forecasting $\vv{u}(t)$ for $t > 0$ based on an initial condition $\vv{u}(0)$ and possibly recent values of $\vv{u}(t)$ for $t < 0$. For notational use in our hybrid scheme (Sec.~\ref{sec:sub_hyb}), we denote integration of the knowledge-based model forward in time by a time duration $\Delta t$ as,
\begin{equation}
\vv{u}_{\mathcal{K}}(t + \Delta t) = \mathcal{K}\left[\vv{u}(t)\right]\approx \vv{u}(t+\Delta t).
\end{equation}
We emphasize that the knowledge-based one-step-ahead predictor $\mathcal{K}$ is imperfect and may have substantial unwanted error. In our test examples in Secs.~\ref{sec:lorenz} and \ref{sec:ks} we consider prediction of continuous-time systems and take the prediction system time step $\Delta t$ to be small compared to the typical time scale over which the continuous-time system changes. We note that while a single prediction time step ($\Delta t$) is small, we are interested in predicting for a large number of time steps.

\subsection{Reservoir-Only Model}\label{sec:sub_res}
For the machine learning approach, we assume the knowledge of $\mathbf{u}(t)$ for times $t$ from $-T$ to $0$. This data will be used to train the machine learning model for the purpose of making predictions of $\vv{u}(t)$ for $t > 0$. In particular we use a reservoir computer, described as follows.
 
A reservoir computer (Fig.~\ref{fig:res_diagram}) is constructed with an artificial high dimensional dynamical system, known as the reservoir whose state is represented by the $D_r$ dimensional vector $\vv{r}(t)$, $D_r \gg M$. We note that ideally the forecasting accuracy of a reservoir-only prediction model increases with $D_r$, but that $D_r$ is typically limited by computational cost considerations. The reservoir is coupled to an input through an Input-to-Reservoir coupling $\hat{\vv{R}}_{\text{in}}\left[\vv{u}(t)\right]$ which maps the $M$-dimensional input vector, $\vv{u}$, at time $t$, to each of the $D_r$ reservoir state variables. The output is defined through a Reservoir-to-Output coupling $\hat{\vv{R}}_{\text{out}}\left[\vv{r}(t), \vv{p}\right]$, where $\vv{p}$ is a large set of \textit{adjustable} parameters.
In the task of prediction of state variables of dynamical systems the reservoir computer is used in two different configurations. One of the configurations we call the `training' phase, and the other one we called the `prediction' phase. In the training phase, the reservoir is configured according to Fig.~\ref{fig:res_diagram} with the switch in the position labeled `Training'. In this phase, the reservoir evolves from $t = -T$ to $t = 0$ according to the equation, 
\begin{equation}\label{eq:input_reservoir}
\vv{r}(t + \Delta t) = \hat{\vv{G}}_R\left[\hat{\vv{R}}_{\text{in}}\left[\vv{u}(t)\right], \vv{r}(t) \right],    
\end{equation}
where the nonlinear function $\hat{\vv{G}}_R$ and the (usually linear) function $\hat{\vv{R}}_{\text{in}}$ depend on the choice of the reservoir implementation. Next, we make a particular choice of the parameters $\vv{p}$ such that the output function $\hat{\vv{R}}_{\text{out}}\left[\vv{r}(t), \vv{p}\right]$ satisfies,
\begin{equation*}
\hat{\vv{R}}_{\text{out}}\left[\vv{r}(t), \vv{p}\right] \approx \vv{u}(t), 
\end{equation*}
for $-T < t \leq 0$. We achieve this by minimizing the error between $\tilde{\vv{u}}_R(t)  = \hat{\vv{R}}_{\text{out}}\left[\vv{r}(t), \vv{p}\right]$ and $\vv{u}(t)$ for $-T < t \leq 0$ using a suitable error metric and optimization algorithm on the adjustable parameter vector $\vv{p}$.

In the prediction phase, for $t\geq 0,$ the switch is placed in position labeled `Prediction' indicated in Fig.~\ref{fig:res_diagram}. The reservoir now evolves autonomously with a feedback loop according to the equation,
\begin{equation}\label{eq:auto_reservoir}
\vv{r}(t+ \Delta t) = \hat{\vv{G}}_R\left[\hat{\vv{R}}_{\text{in}}\left[\tilde{\vv{u}}_R(t)\right], \vv{r}(t)\right],
\end{equation}
where, $\tilde{\vv{u}}_R(t)=\hat{\vv{R}}_{\text{out}}\left[\vv{r}(t), \vv{p}\right]$ is taken as the prediction from this reservoir-only approach. It has been shown~\cite{jaeger2004harnessing} that this procedure can successfully generate a time series $\tilde{\vv{u}}_R(t)$ that approximates the true state $\vv{u}(t)$ for $t > 0$. Thus $\tilde{\vv{u}}_R(t)$ is our reservoir-based prediction of the evolution of $\vv{u}(t)$. If, as assumed henceforth, the dynamical system being predicted is chaotic, the exponential divergence of initial conditions in the dynamical system implies that any prediction scheme will only be able to yield an accurate prediction for a limited amount of time.

\subsection{Hybrid Scheme}\label{sec:sub_hyb}

\begin{figure}[h]
	\centering
	\includegraphics[width=0.5\textwidth]{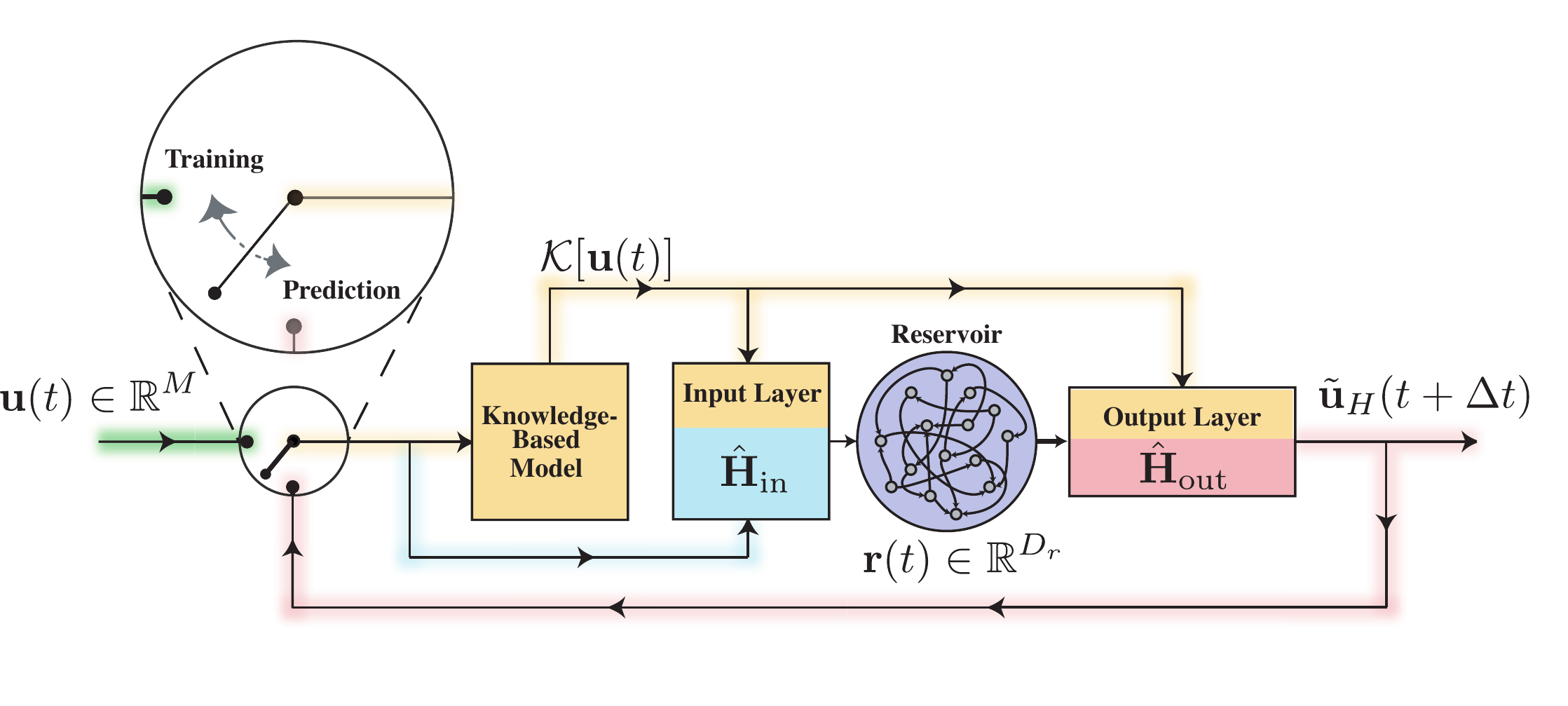}
	\caption{Schematic diagram of the hybrid prediction setup.}
	\label{fig:hybrid_diagram}
\end{figure}

The hybrid approach we propose combines both the knowledge-based model and the reservoir-only model. Our hybrid approach is outlined in the schematic diagram shown in Fig.~\ref{fig:hybrid_diagram}.

As in the reservoir-only model, the hybrid scheme has two phases, the training phase and the prediction phase. In the training phase (with the switch in position labeled `Training' in Fig.~\ref{fig:hybrid_diagram}), the training data $\vv{u}(t)$ from $t = -T$ to $t = 0$ is fed into both the knowledge-based predictor and the reservoir. At each time $t$, the output of the knowledge-based predictor is the one-step ahead prediction $\mathcal{K}\left[\vv{u}(t)\right]$. The reservoir evolves according to the equation

\begin{align}\label{eq:input_hybrid}
\vv{r}(t + \Delta t) &= \hat{\vv{G}}_H\left[\vv{r}(t), \hat{\vv{H}}_{\text{in}}\left[\mathcal{K}\left[\vv{u}(t)\right], \vv{u}(t)\right]\right]
\end{align}
for $-T \leq t \leq 0$, where the (usually linear) function $\hat{\vv{H}}_{\text{in}}$ couples the reservoir network with the inputs to the reservoir, in this case $\vv{u}(t)$ and $\mathcal{K}\left[\vv{u}(t) \right]$. As earlier, we modify a set of adjustable parameters $\vv{p}$ in a predefined output function so that

\begin{align}\label{eq:hybrid_optimization}
\hat{\vv{H}}_{\text{out}}\left[\mathcal{K}\left[\vv{u}(t-\Delta t)\right], \vv{r}(t), \vv{p}\right] \approx \vv{u}(t)
\end{align}
for $-T < t \leq 0$, which is achieved by minimizing the error between the right-hand side and the left-hand side of Eq.~(\ref{eq:hybrid_optimization}), as discussed earlier (Sec. \ref{sec:sub_res}) for the reservoir-only approach. Note that both the knowledge-based model and the reservoir feed into the output layer (Eq.~(\ref{eq:hybrid_optimization}) and Fig.~\ref{fig:hybrid_diagram}) so that the training can be thought of as optimally deciding on how to weight the information from the knowledge-based and reservoir components.

For the prediction phase (the switch is placed in the position labeled `Prediction' in Fig.~\ref{fig:hybrid_diagram}) the feedback loop is closed allowing the system to evolve autonomously. The dynamics of the system will then be given by

\begin{align}\label{eq:auto_hybrid}
\vv{r}(t + \Delta t) &= \hat{\vv{G}}_H\left[ \vv{r}(t), \hat{\vv{H}}_{\text{in}}\left[\mathcal{K}\left[\tilde{\vv{u}}_H(t)\right], \tilde{\vv{u}}_H(t)\right]\right],
\end{align}
where $\tilde{\vv{u}}_H(t) = \hat{\vv{H}}_{\text{out}}\left[\mathcal{K}\left[\tilde{\vv{u}}_H(t - \Delta t)\right],\vv{r}(t), \vv{p}\right]$, is the prediction of the prediction of the hybrid system.

\section{Implementation}\label{sec:implementation}
In this section we provide details of our specific implementation and discuss the prediction performance metrics we use to assess and compare the various prediction schemes. Our implementation of the reservoir computer closely follows Ref.~\cite{jaeger2004harnessing}. Note that, in the reservoir training, no knowledge of the dynamics and details of the reservoir system is used (this contrasts with other machine learning techniques~\cite{goodfellow2016deep}): only the $-T \leq t \leq 0$ training data is used ($\vv{u}(t)$, $\vv{r}(t)$, and, in the case of the hybrid, $\mathcal{K}[\vv{u}(t)]$). This feature implies that reservoir computers, as well as the reservoir-based hybrid are insensitive to the specific reservoir implementation. In this paper, our illustrative implementation of the reservoir computer uses an artificial neural network for the realization of the reservoir. We mention, however, that alternative implementation strategies such as utilizing nonlinear optical devices~\cite{larger2012photonic, antonik2017brain, larger2017high} and Field Programmable Gate Arrays~\cite{haynes2015reservoir} can also be used to construct the reservoir component of our hybrid scheme (Fig.~\ref{fig:hybrid_diagram}) and offer potential advantages, particularly with respect to speed. 

\subsection{Reservoir-Only and Hybrid Implementations}\label{sec:rh_implementation}
Here we consider that the high-dimensional reservoir is implemented by a large, low degree Erd\H{o}s-R\`{e}nyi network of $D_r$ nonlinear, neuron-like units in which the network is described by an adjacency matrix $\vv{A}$ (we stress that the following implementations are somewhat arbitrary, and are intended as illustrating typical results that might be expected). The network is constructed to have an average degree denoted by $\langle d \rangle$, and the nonzero elements of $\vv{A}$, representing the edge weights in the network, are initially chosen independently from the uniform distribution over the interval $[-1,1]$. All the edge weights in the network are then uniformly scaled via multiplication of the adjacency matrix by a constant factor to set the largest magnitude eigenvalue of the matrix to a quantity $\rho$, which is called the `spectral radius' of $\vv{A}$. The state of the reservoir, given by the vector $\vv{r}(t)$, consists of the components $r_j$ for $1\leq j \leq D_r$ where $r_j(t)$ denotes the scalar state of the $j^{\text{th}}$ node in the network. 
When evaluating prediction based purely on a reservoir system alone, the reservoir is coupled to the $M$ dimensional input through a $D_r \times M$ dimensional matrix $\vv{W}_{\text{in}}$, such that in Eq.~(\ref{eq:input_reservoir}) $\hat{\vv{R}}_{\text{in}}\left[\vv{u}(t) \right] = \vv{W}_{\text{in}}\vv{u}(t)$, and each row of the matrix $\vv{W}_{\text{in}}$ has exactly one randomly chosen nonzero element. Each nonzero element of the matrix is independently chosen from the uniform distribution on the interval $[-\sigma,\sigma]$. We choose the hyperbolic tangent function for the form of the nonlinearity at the nodes, so that the specific training phase equation for our reservoir setup corresponding to Eq.~(\ref{eq:input_reservoir}) is
\begin{align}
    \vv{r}(t + \Delta t) &= \tanh\left[\vv{A}\vv{r}(t) + \vv{W}_{\text{in}}\vv{u}(t)\right],
\end{align}
where the hyperbolic tangent applied on a vector is defined as the vector whose components are the hyperbolic tangent function acting on each element of the argument vector individually. 

We choose the form of the output function to be $\hat{\vv{R}}_{\text{out}}(\vv{r}, \vv{p}) = \vv{W}_{\text{out}}\vv{r}^\star$, in which the output parameters (previously symbolically represented by $\vv{p}$) will henceforth be take to be the elements of the matrix $\vv{W}_{\text{out}}$, and the vector $\vv{r}^\star$ is defined such that $r^\star_j$ equals $r_j$ for odd $j$, and equals $r_j^2$ for even $j$ (it was empirically found that this choice of $\vv{r}^\star$ works well for our examples in both Sec.~\ref{sec:lorenz} and Sec.~\ref{sec:ks}, see also~\cite{lu2017reservoir, pathakmodel}). We run the reservoir for $-T \leq t \leq 0$ with the switch in Fig.~\ref{fig:res_diagram} in the `Training' position. We then minimize $\sum_{m=1}^{T/\Delta t}~\lVert~\vv{u}(-m\Delta t)-\tilde{\vv{u}}_R(-m\Delta t) \rVert^2$ with respect to $\vv{W}_{\text{out}}$, where $\tilde{\vv{u}}_R$ is now $\vv{W}_{\text{out}} \vv{r}^\star$. Since $\tilde{\vv{u}}_R$ depends linearly on the elements of $\vv{W}_{\text{out}}$, this minimization is a standard linear regression problem, and we use Tikhonov regularized linear regression \cite{tikhonov1977solutions}. We denote the regularization parameter in the regression by $\beta$ and employ a small positive value of $\beta$ to prevent over fitting of the training data.

Once the output parameters (the matrix elements of $\vv{W}_{\text{out}}$) are determined, we run the system in the configuration depicted in Fig. \ref{fig:res_diagram} with the switch in the `Prediction' position according to the equations,
\begin{align}
\tilde{\vv{u}}_R(t) &= \vv{W}_{\text{out}}\vv{r}^\star(t)\\
\vv{r}(t + \Delta t) &= \tanh\left[\vv{A}\vv{r}(t) + \vv{W}_{\text{in}}\tilde{\vv{u}}_R(t) \right],
\end{align}
corresponding to Eq.~(\ref{eq:auto_reservoir}). Here $\tilde{\vv{u}}_R(t)$ denotes the prediction of $\vv{u}(t)$ for $t > 0$ made by the reservoir-only model.

Next, we describe the implementation of the hybrid prediction scheme. The reservoir component of our hybrid scheme is implemented in the same fashion as in the reservoir-only model given above. In the training phase for $-T<t\leq 0$, when the switch in Fig.~\ref{fig:hybrid_diagram} is in the `Training' position, the specific form of Eq.~(\ref{eq:input_hybrid}) used is given by
\begin{align}
\vv{r}(t + \Delta t) &=  \tanh\left[\vv{A}\vv{r}(t) + \vv{W}_{\text{in}} \left( \begin{matrix} \mathcal{K}\left[\vv{u}(t)\right] \\ \vv{u}(t) \end{matrix} \right) \right]
\end{align}
As earlier, we choose the matrix $\vv{W}_{\text{in}}$ (which is now $D_r\times(2 M)$ dimensional) to have exactly one nonzero element in each row. The nonzero elements are independently chosen from the uniform distribution on the interval $[-\sigma,\sigma]$. Each nonzero element can be interpreted to correspond to a connection to a particular reservoir node. These nonzero elements are randomly chosen such that a fraction $\gamma$ of the reservoir nodes are connected exclusively to the raw input $\vv{u}(t)$ and the remaining fraction are connected exclusively to the the output of the model based predictor $\mathcal{K}[\vv{u}(t)]$.

Similar to the reservoir-only case, we choose the form of the output function to be 
\begin{align}
\hat{\vv{H}}_{\text{out}}\left[\mathcal{K}\left[ \vv{u}(t - \Delta t) \right], \vv{r}(t), \vv{p} \right] =  \vv{W}_{\text{out}}\left(\begin{matrix} \mathcal{K}\left[\vv{u}(t-\Delta t)\right] \\ \vv{r}^\star(t) \end{matrix} \right),
\end{align}
Where, as in the reservoir-only case, $\vv{W}_{\text{out}}$ now plays the role of $\vv{p}$. Again, as in the reservoir-only case, $\vv{W}_{\text{out}}$ is determined via Tikhonov regularized regression.

In the prediction phase for $t>0$, when the switch in Fig. \ref{fig:hybrid_diagram} is in position labeled `Prediction', the input $\vv{u}(t)$ is replaced by the output at the previous time step and the equation analogous to Eq.~(\ref{eq:auto_hybrid}) is given by,
\begin{align}
\tilde{\vv{u}}_H(t) &= \vv{W}_{\text{out}}\left(\begin{matrix} \mathcal{K}\left[\vv{u}(t) \right] \\ \vv{r}^\star(t)\end{matrix}\right),\\
\vv{r}(t + \Delta t) &=  \tanh\left[\vv{A}\vv{r}(t) + \vv{W}_{\text{in}} \left( \begin{matrix} \mathcal{K}\left[\tilde{\vv{u}}_H \right] \\ \vv{\tilde{u}}_H \end{matrix} \right)\right].
\end{align}
The vector time series $\tilde{\vv{u}}_H(t)$ denotes the prediction of $\vv{u}(t)$ for $t > 0$ made by our hybrid scheme.

\subsection{Training Reusability}\label{sec:reusability}

In the prediction phase, $t > 0$, chaos combined with a small initial condition error, $\lVert \tilde{\vv{u}}(0) - \vv{u}(0) \rVert \ll \lVert \vv{u}(0) \rVert$, and imperfect reproduction of the true system dynamics by the prediction method lead to a roughly exponential increase of the prediction error $\lVert \tilde{\vv{u}}(t)-\vv{u}(t) \rVert$ as the prediction time $t$ increases. Eventually, the prediction error becomes unacceptably large. By choosing a value for the largest acceptable prediction error, one can define a ``valid time" $t_v$ for a particular trial. As our examples in the following sections show, $t_v$ is typically much less than the necessary duration $T$ of the training data required for either reservoir-only prediction or for prediction by our hybrid scheme. However, it is important to point out that the reservoir and hybrid schemes have the property of \textit{training reusability}. That is, once the output parameters $\vv{p}$ (or $\vv{W}_{\text{out}}$) are obtained using the training data in $-T\leq t \leq 0$, the same $\vv{p}$ can be used over and over again for subsequent predictions, without retraining $\vv{p}$. For example, say that we now desire another prediction starting at some later time $t_0 > 0$. In order to do this, the reservoir system (Fig.~\ref{fig:res_diagram}) or the hybrid system (Fig.~\ref{fig:hybrid_diagram}) \textit{with the predetermined} $\vv{p}$, is first run with the switch in the `Training' position for a time, $(t_0 - \xi) < t < t_0$. This is done in order to resynchronize the reservoir to the dynamics of the true system, so that the time $t = t_0$ prediction system output, $\tilde{\vv{u}}(t_0)$, is brought very close to the true value, $\vv{u}(t_0)$, of the process to be predicted for $t > t_0$ (i.e., the reservoir state is resynchronized to the dynamics of the chaotic process that is to be predicted). Then, at time $t = t_0$, the switch (Figs.~\ref{fig:res_diagram} and~\ref{fig:hybrid_diagram}) is moved to the `Prediction' position, and the output $\tilde{\vv{u}}_R$ or $\tilde{\vv{u}}_H$ is taken as the prediction for $t > t_0$. We find that with $\vv{p}$ predetermined, the time required for re-synchronization $\xi$ turns out to be very small compared to $t_v$, which is in turn small compared to the training time $T$.

\subsection{Assessments of Prediction Methods}\label{sec:assessment}

We wish to compare the effectiveness of different prediction schemes (knowledge-based, reservoir-only, or hybrid). As previously mentioned, for
each independent prediction, we quantify the duration of accurate prediction with the corresponding ``valid time", denoted $t_v$, defined as the elapsed time before the normalized error $E(t)$ first exceeds some value $f$, $0 <f <1$, $E(t_v)=f$, where
\begin{equation}
E(t) = \frac{||\mathbf{u}(t) - \mathbf{\widetilde{u}}(t)||}{\langle ||\mathbf{u}(t)||^2 \rangle^{1/2}} ,
\label{eq:rms}
\end{equation}
and the symbol $\tilde{\vv{u}}(t)$ now stands for the prediction [either $\tilde{\vv{u}}_{\mathcal{K}}(t), \tilde{\vv{u}}_{R}(t)$ or $\tilde{\vv{u}}_{H}(t)$ as obtained by either of the three prediction methods (knowledge-based, reservoir-based, or hybrid)]. 

In what follows we use $f = 0.4$. We test all three methods on 20 disjoint time intervals of length $\tau$ in a long run of the true dynamical system. For each prediction method, we evaluate the valid time over many independent prediction trials. Further, for the reservoir-only prediction and the hybrid schemes, we use 32 different random realizations of $\vv{A}$ and $\vv{W}_{\text{in}}$, for each of which we separately determine the training output parameters $\vv{W}_{\text{out}}$; then we predict on each of the 20 time intervals for each such random realization, taking advantage of training reusability (Sec.~\ref{sec:reusability}). Thus, there are a total of 640 different trials for the reservoir-only and hybrid system methods, and 20 trials for the knowledge-based method. We use the median valid time across all such trials as a measure of the quality of prediction of the corresponding scheme, and the span between the first and third quartiles of the $t_v$ values as a measure of variation in this metric of the prediction quality. 
  
\section{Lorenz system}\label{sec:lorenz}
The Lorenz system~\cite{lorenz1963deterministic} is described by the equations,
	\begin{align}\label{eq:lorenz}
		\frac{dx}{dt} &= -ax + ay, \nonumber\\
		\frac{dy}{dt} &= bx - y - xz,\\
		\frac{dz}{dt} &= -cz + xy, \nonumber
	\end{align}

For our ``true" dynamical system, we use $a = 10$, $b=28$, $c=8/3$ and we  generate a long trajectory with $\Delta t = 0.1$. For our knowledge-based predictor, we use an `imperfect' version of the Lorenz equations to represent an approximate, imperfect model that might be encountered in a real life situation. Our imperfect model differs from the true Lorenz system given in Eq.~(\ref{eq:lorenz}) only via a change in the system parameter $b$ in Eq.~(\ref{eq:lorenz}) to $b(1+\epsilon)$. The error parameter $\epsilon$ is thus a dimensionless quantification of the discrepancy between our knowledge-based predictor and the `true' Lorenz system. We emphasize that, although we simulate model error by a shift of the parameter $b$, we view this to represent a general model error of \textit{unknown form}. This is reflected by the fact that our reservoir and hybrid methods do not incorporate knowledge that the system error in our experiments results from an imperfect parameter value of a system with Lorenz form. 

Next, for the reservoir computing component of the hybrid scheme, we construct a network-based reservoir as discussed in Sec.~\ref{sec:sub_res} for various reservoir sizes $D_r$ and with the parameters listed in Table~\ref{tab:params_lorenz}.

\begin{table}[h]
\centering
\begin{tabular}{ c  r || c  r }
	Parameter & Value & Parameter & Value\\
	\hline
	$\rho$ & $0.4$ & $T$ & $100$\\
 	$\langle d \rangle$ & $3$ & $\gamma$ & $0.5$   \\
	$\sigma$ & $0.15$ & $\tau$ & $250$ \\
	$\Delta t$ & $0.1$ & $\xi$ & $10$\\
\end{tabular}
\caption{Reservoir parameters $\rho$, $\langle d \rangle$, $\sigma$, $\Delta t$, training time $T$, hybrid parameter $\gamma$, and evaluation parameters $\tau$, $\xi$ for the Lorenz system prediction.}
\label{tab:params_lorenz}
\end{table}

\begin{figure}[h]
	\centering
	\includegraphics[width=0.4\textwidth]{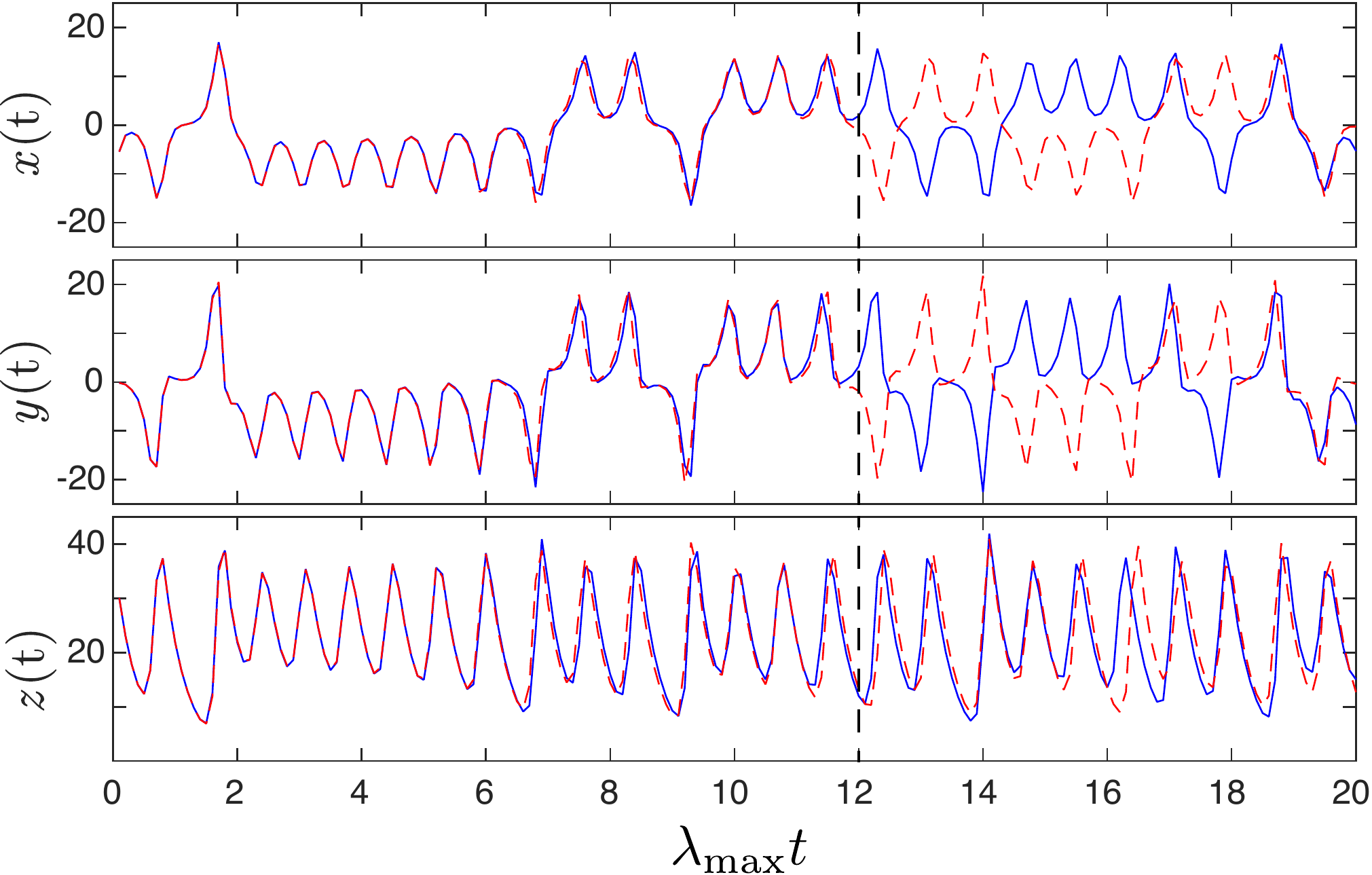}
	\caption{Prediction of the Lorenz system using the hybrid prediction setup. The blue line shows the true state of the Lorenz system and the red dashed line shows the prediction. Prediction begins at $t = 0$. The vertical black dashed line marks the point where this prediction is no longer considered valid by the valid time metric with $f = 0.4$. The error in the approximate model used in the knowledge-based component of the hybrid scheme is $\epsilon = 0.05$.}
	\label{fig:hybrid_pred}
\end{figure}

\begin{figure}[h]
	\centering
	\includegraphics[width=0.4\textwidth]{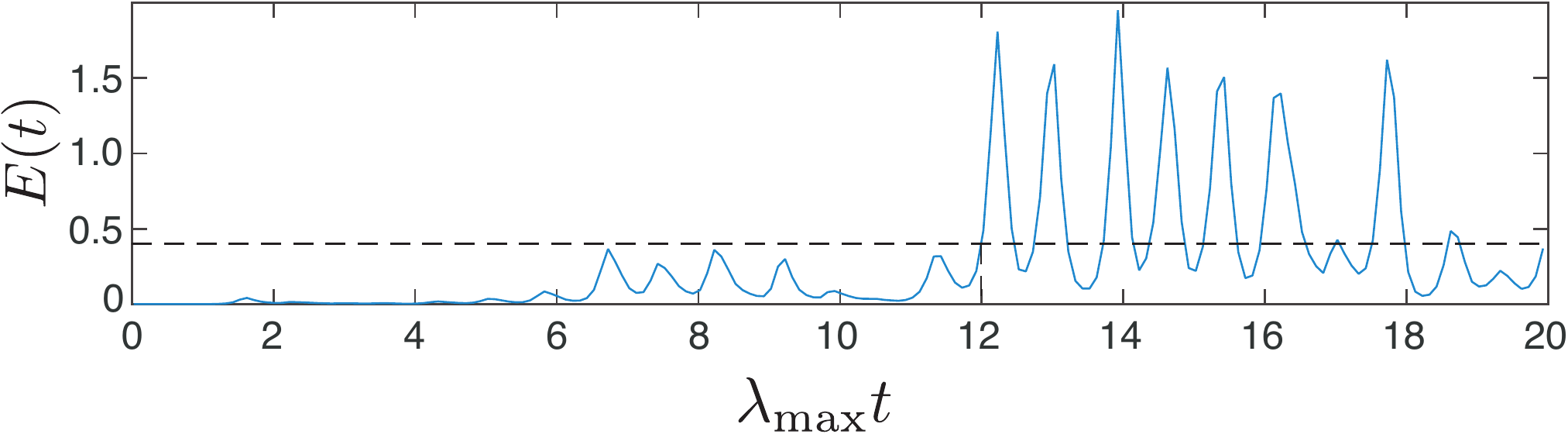}
	\caption{Normalized error $E(t)$ versus time of the Lorenz prediction trial shown in Fig.~\ref{fig:hybrid_pred}. The prediction error remains below the defined threshold ($E(t) < 0.4$) for about 12 Lyapunov times.}
\label{fig:lrz_valid}
\end{figure}

Figure \ref{fig:hybrid_pred} shows an illustrative example of one prediction trial using the hybrid method. The horizontal axis is the time in units of the Lyapunov time $\lambda^{-1}_{\text{max}}$, where $\lambda_{\text{max}}$ denotes the largest Lyapunov exponent of the system, Eqs.~(\ref{eq:lorenz}). The vertical dashed lines in Fig.~\ref{fig:hybrid_pred} indicate the valid time $t_v$ (Sec.~\ref{sec:assessment}) at which $E(t)$ (Eq.~(\ref{eq:rms})) first reaches the value $f = 0.4$. The valid time determination for this example with $\epsilon = 0.05$ and $D_r = 500$ is illustrated in Fig.~\ref{fig:lrz_valid}. Notice that we get low prediction error for about 10 Lyapunov times. 

\begin{figure}[h]
	\centering
	\includegraphics[width=0.4\textwidth]{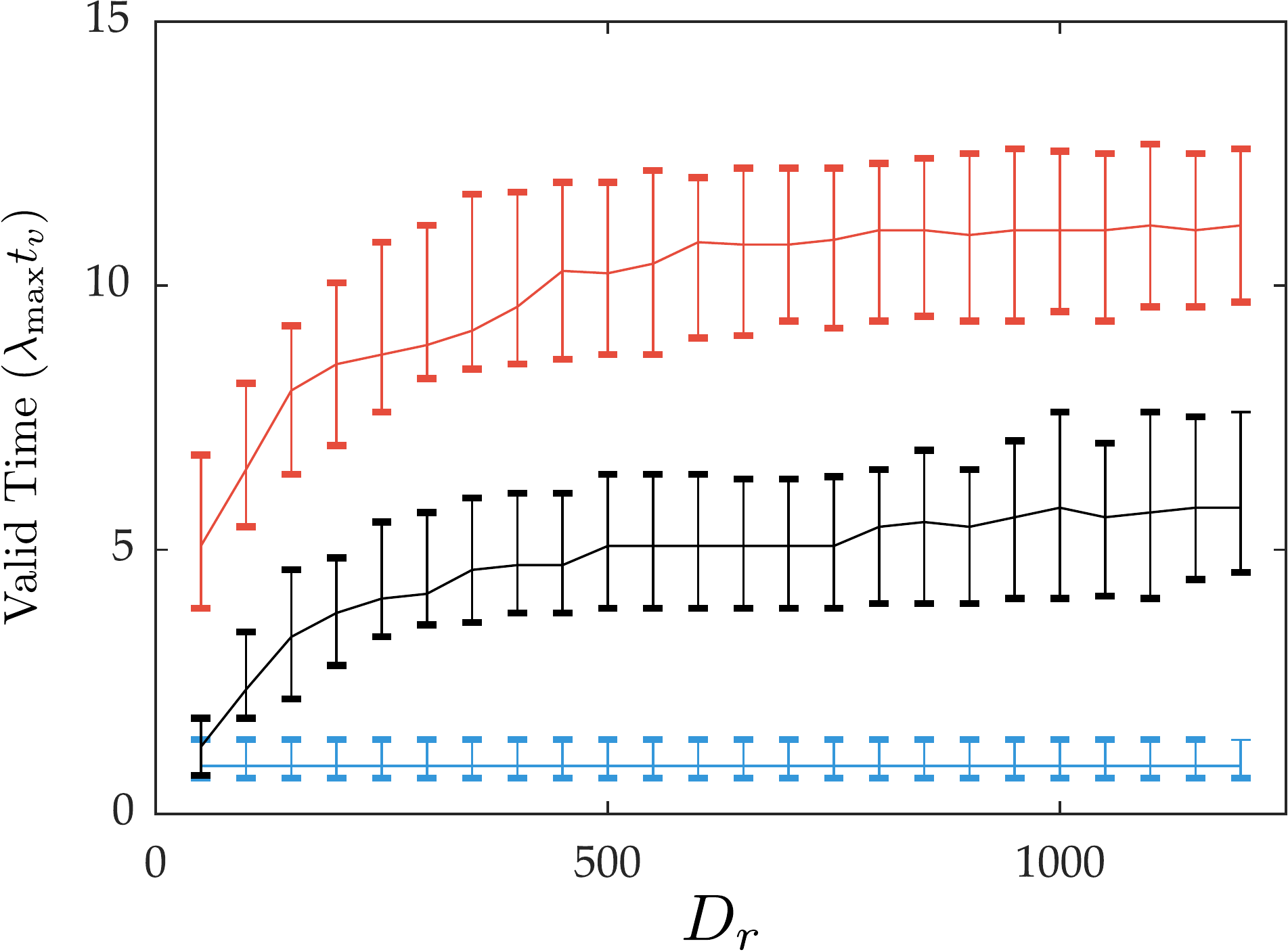}
	\caption{Reservoir size ($D_r$) dependence of the median valid time using the hybrid prediction scheme (red upper plot), the reservoir-only (black middle plot) and the knowledge-based model only methods. The model error is fixed at $\epsilon = 0.05$. Since the knowledge based model (blue) does not depend on $D_r$, its plot is a horizontal line. Error bars span the range between the $1^{\text{st}}$ and $3^{\text{rd}}$ quartiles of the trials.}
	\label{fig:res_size}
\end{figure}

The red upper curve in Fig.~\ref{fig:res_size} shows the dependence on reservoir size $D_r$ of results for the median valid time (in units of Lyapunov time, $\lambda_{\text{max}}t$, and with $f = 0.4$) of the predictions from a hybrid scheme using a reservoir system combined with our imperfect model with an error parameter of $\epsilon = 0.05$. The error bars span the first and third quartiles of our trials which are generated as described in Sec.~\ref{sec:assessment}. The black middle curve in Fig.~\ref{fig:res_size} shows the corresponding results for predictions using the reservoir-only model. The blue lower curve in Fig.~\ref{fig:res_size} shows the result for prediction using only the $\epsilon = 0.05$ imperfect knowledge-based model (since this result does not depend on $D_r$, the blue curve is horizontal and the error bars are the same at each value of $D_r$). Note that, even though the knowledge-based prediction alone is very bad, when used in the hybrid, it results in a large prediction improvement relative to the reservoir-only prediction. Moreover, this improvement is seen for all values of the reservoir sizes tested. Note also that the valid time for the hybrid with a reservoir size of $D_r = 50$ is comparable to the valid time for a reservoir-only scheme at $D_r = 500$. This suggests that our hybrid method can substantially reduce reservoir computational expense even with a knowledge-based model that has low predictive power on its own.

\begin{figure}[h]
	\centering
	\includegraphics[width=0.4\textwidth]{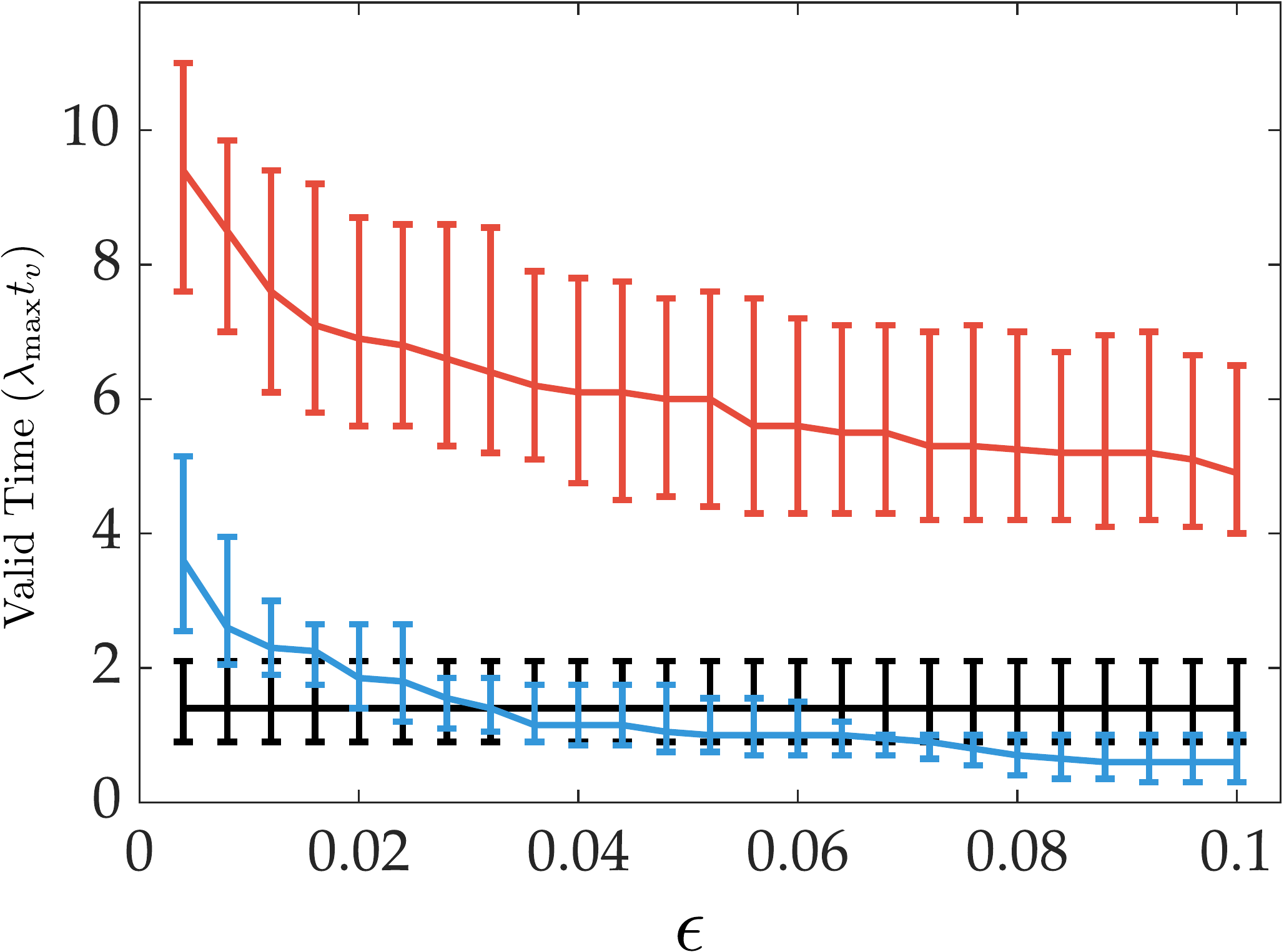}
	\caption{Valid times for different values of model error ($\epsilon$) with $f = 0.4$. The reservoir size is fixed at $D_r=50$. Plotted points represent the median and error bars span the range between the $1^{\text{st}}$ and $3^{\text{rd}}$ quartiles. The meaning of the colors is the same as in Fig.~\ref{fig:res_size}. Since the reservoir only scheme (black) does not depend on $\epsilon$, its plot is a horizontal line. Similar to Fig.~\ref{fig:res_size}, the small reservoir alone cannot predict well for a long time, but the hybrid model, which combines the inaccurate knowledge-based model and the small reservoir performs well across a broad range of $\epsilon$.}
	\label{b_error}
\end{figure}

Fig.~\ref{b_error} shows the dependence of prediction performance on the model error $\epsilon$ with the reservoir size held fixed at $D_r = 50$. For the wide range of the error $\epsilon$ we have tested, the hybrid performance is much better than either its knowledge-based component alone or reservoir-only component. Figures~\ref{fig:res_size} and \ref{b_error}, taken together, suggest the potential robustness of the utility of the hybrid approach.

\section{Kuramoto-Sivashinsky equations}\label{sec:ks}

\begin{figure*}[h]
	\centering
	\includegraphics[width = 0.4\textwidth]{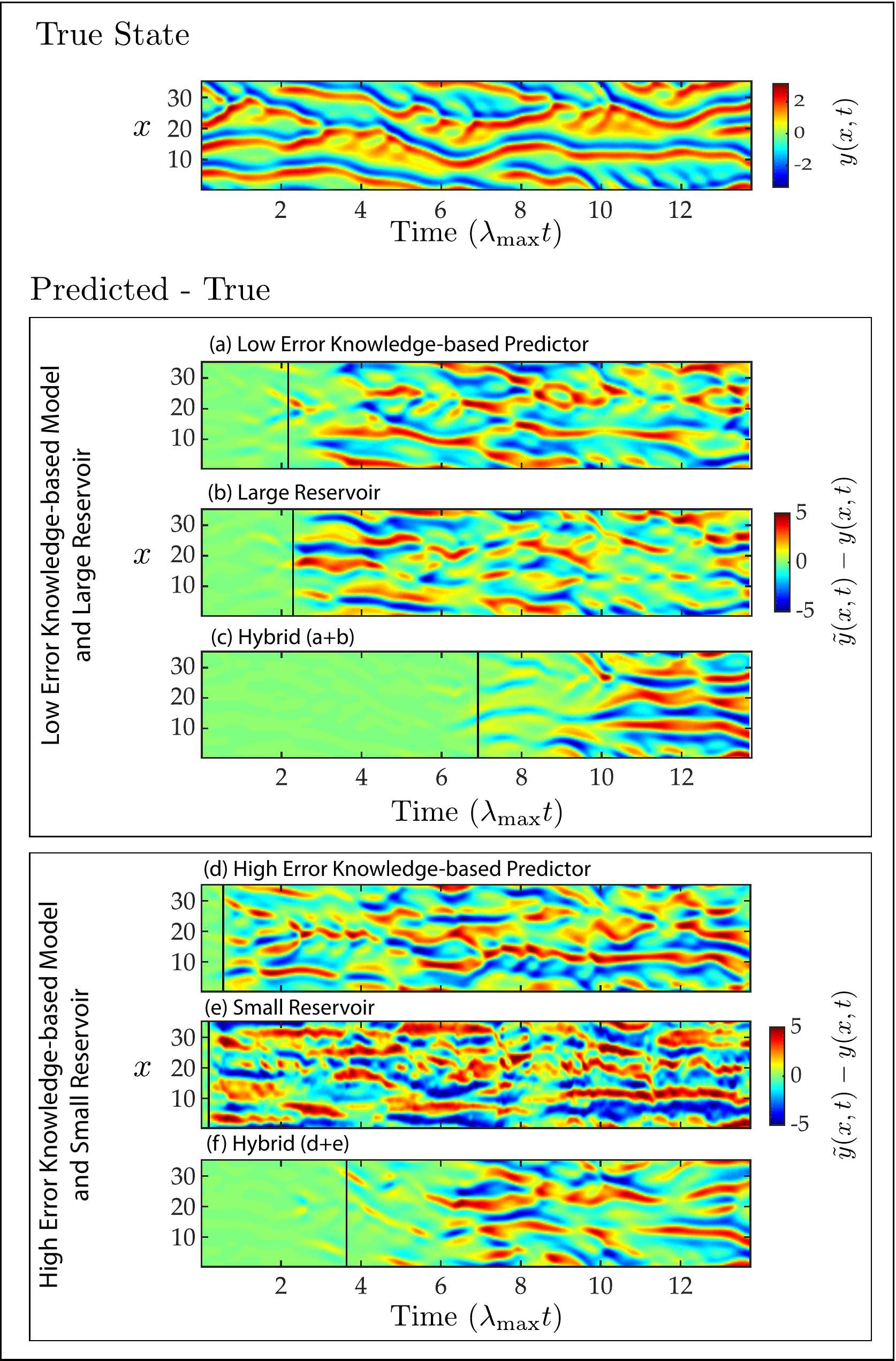}
	\caption{The topmost panel shows the true solution of the KS equation (Eq.~(\ref{eq:KS})). Each of the six panels labeled (a) through (f) shows the difference between the true state of the KS system and the prediction made by a specific prediction scheme. The three panels (a), (b), and (c) respectively show the results for a low error knowledge-based model ($\epsilon = 0.01$), a reservoir-only prediction scheme with a large reservoir ($D_r = 8000$), and the hybrid scheme composed of ($D_r =8000$, $\epsilon = 0.01$). The three panels, (d), (e), and (f) respectively show the corresponding results for a highly imperfect knowledge-based model ($\epsilon = 0.1$), a reservoir-only prediction scheme using a small reservoir ($D_r = 500$), and the hybrid scheme with ($D_r = 500$, $\epsilon = 0.1$).}
	\label{fig:ksplots}
\end{figure*}

\begin{figure*}[h]
	\centering
	\includegraphics[width = 0.4\textwidth]{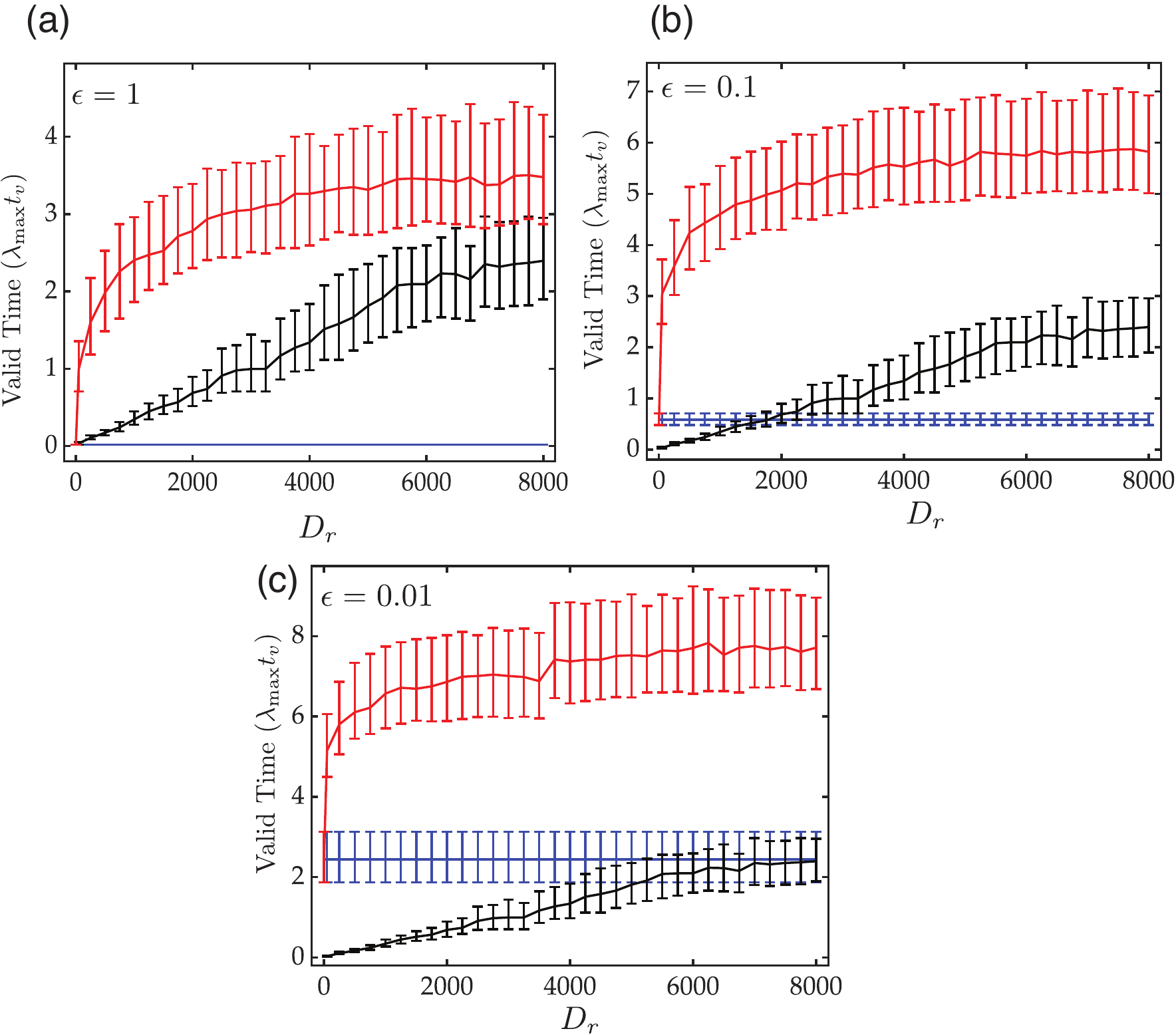}
	\caption{Each of the three panels (a), (b), and (c) shows a comparison of the KS system prediction performance of the reservoir-only scheme (black), the knowledge-based model (blue) and the hybrid scheme (red). The median valid time in Lyapunov units ($\lambda_{\text{max}}t_v$) is plotted against the size of the reservoir used in the hybrid scheme and the reservoir-only scheme. Since the knowledge-based model does not use a reservoir, its valid time does not vary with the reservoir size. The error in the knowledge-based model is $\epsilon = 1$ in panel (a), $\epsilon = 0.1$ in panel (b) and $\epsilon = 0.01$ in panel (c).}
	\label{fig:ks_valid_time}
\end{figure*}

In this example, we test how well our hybrid method, using an inaccurate knowledge-based model combined with a relatively small reservoir, can predict systems that exhibit high dimensional spatiotemporal chaos. Specifically, we use simulated data from the one-dimensional Kuramoto-Sivashinsky (KS) equation for $y(x,t)$,
\begin{equation}
y_t = -yy_x-y_{xx}-y_{xxxx}
\label{eq:KS}
\end{equation}
Our simulation calculates $y(x,t)$ on a uniformly spaced grid with spatially periodic boundary conditions such that $y(x,t) = y(x+L,t)$, with a periodicity length of $L=35$, a grid size of $Q = 64$ grid points (giving a intergrid spacing of $\Delta x =\frac{L}{Q} \approx 0.547$), and a sampling time of $\Delta t = 0.25$. For these parameters we found that the maximum Lyapunov exponent, $\lambda_{\text{max}}$, is positive ($\lambda_{\text{max}} \approx 0.07$), indicating that this system is chaotic. We define a vector of $y(x,t)$ values at each grid point as the input to each of our predictors:
\begin{equation}
\mathbf{u}(t) = \left[y\left(\frac{L}{Q},t\right),
y\left(\frac{2L}{Q},t\right),\dots,y\left(L,t\right)\right]^T.
\end{equation}

For our approximate knowledge-based predictor, we use the same simulation method as the original Kuramoto-Sivashinsky equations with an error parameter $\epsilon$ added to the coefficient of the second derivative term as follows,
\begin{equation}\label{eq:KS_error}
y_t = -yy_x-(1+\epsilon)y_{xx}-y_{xxxx}.
\end{equation}
For sufficiently small $\epsilon$, Eq.~(\ref{eq:KS_error}) corresponds to a very accurate knowledge-based model of the true KS system, which becomes less and less accurate as $\epsilon$ is increased.

Illustrations of our main result are shown in Figs.~\ref{fig:ksplots} and \ref{fig:ks_valid_time}, where we use the parameters in Table \ref{tab:params_ks}. In the top panel of Fig.~\ref{fig:ksplots}, we plot a computed solution of Eq.~(\ref{eq:KS}) which we regard as the true dynamics of a system to be predicted; the spatial coordinate $x \in [0, L]$ is plotted vertically, the time in Lyapunov units ($\lambda_{\text{max}}t$) is plotted horizontally, and the value of $y(x,t)$ is color coded with the most positive and most negative $y$ values indicated by red and blue, respectively. Below this top panel are six panels labeled (a-f) in which the color coded quantity is the prediction error $\tilde{y}(x,t) - y(x,t)$ of different predictions $\tilde{y}(x,t)$. In panels (a), (b) and (c), we consider a case ($\epsilon = 0.01$, $D_r = 8000$) where both the knowledge-based model (panel (a)) and the reservoir-only predictor (panel (b)) are fairly accurate; panel (c) shows the hybrid prediction error. In panels (d), (e), and (f), we consider a different case ($\epsilon = 0.1$, $D_r = 500$) where both the knowledge-based model (panel (d)) and the reservoir-only predictor (panel (e)) are relatively inaccurate; panel (f) shows the hybrid prediction error. In our color coding, low prediction error corresponds to the green color. The vertical solid lines denote the valid times for this run with $f = 0.4$.

\begin{table}[h]
\centering
\begin{tabular}{ c  r || c  r }
	Parameter & Value & Parameter & Value\\
	\hline
	$\rho$ & $0.4$ & $T$ & $5000$\\
 	$\langle d \rangle$ & $3$ & $\gamma$ & $0.5$   \\
	$\sigma$ & $1.0$ & $\tau$ & $250$ \\
	$\Delta t$ & $0.25$ & $\xi$ & $10$\\
\end{tabular}
\caption{Reservoir parameters $\rho$, $\langle d \rangle$, $\sigma$, $\Delta t$, training time $T$, hybrid parameter $\gamma$, and evaluation parameters $\tau$, $\xi$ for the KS system prediction.}
\label{tab:params_ks}
\end{table}

We note from Figs.~\ref{fig:ksplots}(a,b,c), that even when the knowledge-based model prediction is valid for about as long as the reservoir-only prediction, our hybrid scheme can significantly outperform both its components. Additionally, as in our Lorenz system example (Fig.~\ref{b_error} for $\epsilon \gtrsim 0.2$) we see from Figs.~\ref{fig:ksplots}(d,e,f) that in the parameter regimes where the KS reservoir-only model and knowledge-based model both show very poor performance, the hybrid of these low performing methods can still predict for a longer time than a much more computationally expensive reservoir-only implementation (Fig.~\ref{fig:ksplots}(b)).

This latter remarkable result is reinforced by Fig.~\ref{fig:ks_valid_time}(a), which shows that even for very large error, $\epsilon=1$, such that the model is totally ineffective, the hybrid of these two methods is able to predict for a significant amount of time using a relatively small reservoir. This implies that a non-viable model can be made viable via the addition of a reservoir component of modest size. Further Figs.~\ref{fig:ks_valid_time}(b,c) show that even if one has a model that can outperform the reservoir prediction, as is the case for $\epsilon = 0.01$ for most reservoir sizes, one can still benefit from a reservoir using our hybrid technique.

\section{conclusions}
In this paper we present a method for the prediction of chaotic dynamical systems that hybridizes reservoir computing and knowledge-based prediction. Our main results are:

\begin{enumerate}
\item Our hybrid technique consistently outperforms its component reservoir-only or knowledge-based model prediction methods in the duration of its ability to accurately predict, for both the Lorenz system and the spatiotemporal chaotic Kuramoto-Sivashinsky equations.
\item Our hybrid technique robustly yields improved performance even when the reservoir-only predictor and the knowledge-based model are so flawed that they do not make accurate predictions on their own.
\item Even when the knowledge-based model used in the hybrid is significantly flawed, the hybrid technique can, at small reservoir sizes, make predictions comparable to those made by much larger reservoir-only models, which can be used to save computational resources.
\item Both the hybrid scheme and the reservoir-only model have the property of ``training reusability'' (Sec.~\ref{sec:reusability}), meaning that once trained, they can make any number of subsequent predictions (without retraining each time) by preceding each such prediction with a short run in the training configuration (see Figs.~\ref{fig:res_diagram} and \ref{fig:hybrid_diagram}) in order to resynchronize the reservoir dynamics with the dynamics to be predicted.
\end{enumerate}

\section{Acknowledgment}
This work was supported by ARO (W911NF-12-1-0101), NSF (PHY-1461089) and DARPA.

\end{document}